# CNN Model & Tuning for Global Road Damage Detection

Rahul Vishwakarma
*Big Data Analytics & Solutions Lab*
*Hitachi America Ltd. Research & Development*
Santa Clara, CA, USA
rahul.vishwakarma@hal.hitachi.com

Ravigopal Vennelakanti
*Big Data Analytics & Solutions Lab*
*Hitachi America Ltd. Research & Development*
Santa Clara, CA, USA
ravigopal.vennelakanti@hal.hitachi.com

*Abstract*—This paper provides a report on our solution including model selection, tuning strategy and results obtained for Global Road Damage Detection Challenge. This Big Data Cup Challenge was held as a part of IEEE International Conference on Big Data 2020. We assess single and multi-stage network architectures for object detection and provide a benchmark using popular state-of-the-art open-source PyTorch frameworks like Detectron2 and Yolov5. Data preparation for provided Road Damage training dataset, captured using smartphone camera from Czech, India and Japan is discussed. We studied the effect of training on a per country basis with respect to a single generalizable model. We briefly describe the tuning strategy for the experiments conducted on two-stage Faster R-CNN with Deep Residual Network (Resnet) and Feature Pyramid Network (FPN) backbone. Additionally, we compare this to a one-stage Yolov5 model with Cross Stage Partial Network (CSPNet) backbone. We show a mean F1 score of 0.542 on Test2 and 0.536 on Test1 datasets using a multi-stage Faster R-CNN model, with Resnet-50 and Resnet-101 backbones respectively. This shows the generalizability of the Resnet-50 model when compared to its more complex counterparts. Experiments were conducted using Google Colab having K80 and a Linux PC with 1080Ti, NVIDIA consumer grade GPU. A PyTorch based Detectron2 code to pre-process, train, test and submit the Avg F1 score to is made available at https://github.com/vishwakarmarhl/rdd2020

## I. Introduction

Roads are a critical mobility infrastructure asset that requires condition assessment and monitoring. This was traditionally done by manual survey and expensive inspection methods. The high cost and issues in existing practices like manual labor, specialized inspection equipment's, subject knowledge and logistical delays in assessment are prohibitive. To address this issue, automated image processing from an off-the-shelf smartphone camera has shown to be increasingly effective in visual damage detection [1].

Deep learning based Object detection and localization techniques have shown immense progress in the last decade. Convolutional Neural Network (CNN) using supervised learning approach is applied to image recognition domain that are difficult to model using conventional methods. ImageNet [3] challenge accelerated the object detection task and by year 2015 exceeded human ability. The objective of this work is to assess object detection methods and run experiments to train a damage detection model with the most accurate and generalizable architecture. We achieve the following in this paper.

- Pre-processing to achieve accurate detections
- Train a generalizable model that can be transferred across countries. Contrast it with a dedicated model per country approach.
- Experiment and evaluate single and multi-stage object detectors for accurate detection
- Assess the progress, hyper-parameters and accuracy of various models

The following section describes the data, experiments and analysis to achieve the reported score.

## II. Dataset

The Road Damage Dataset 2020 [2] was curated and annotated for automated inspection. This multi-country dataset is released as a part of IEEE Big Data Cup Challenge [23]. The task is to detect road damages at a global scale and report the performance on Test 1 and Test 2 datasets.

TABLE I. Road Damage type Definitions

| *Damage Type* | | | *Detail* | *Name* |
|---|---|---|---|---|
| Crack | Linear Crack | Longitudinal | Wheel mark part | **D00** |
| | | | Construction Joint part | D01 |
| | | Lateral | Equal Interval | **D10** |
| | | | Construction joint part | D11 |
| | Alligator Crack | | Partial pavement, overall pavement | **D20** |
| Other Corruption | | | Rutting, bump, pothole, separation | **D40** |
| | | | Cross walk blur | D43 |
| | | | White line blur | D44 |

Source Road Maintenance and Repair Guidebook, JRA (2013) [22] in Japan

The damages vary across countries. To generalize the damage category detection in Table I, classes considered for the analysis are; D00: *Longitudinal Crack*, D10: *Transverse Crack*, D20: *Alligator Crack*, D40: *Pothole*.

Test 1 and Test 2 data is provided by the challenge [23] committee for evaluation and submission. Upon submission an Average F1 score is added to the private leaderboard as well as a public leaderboard if it exceeds all the previous scores in our private leaderboard.

### A. Global Road Damage Dataset

The latest dataset is collected from Czech Republic and India in addition to what was made available by GIS Association of Japan. The 2020 dataset provides training images of size 600x600 with damages as a bounding box with associated damage class. Class labels and bounding box coordinates, defined by four numbers ($x_{min}$, $y_{min}$, $x_{max}$, $y_{max}$), are stored in the XML format as per PASCAL VOC [12].

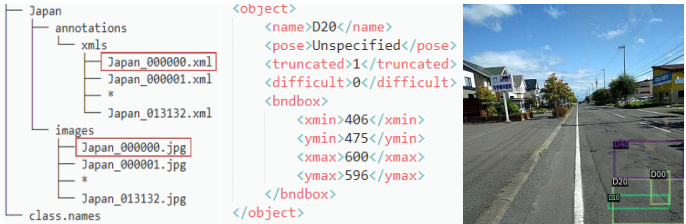

Fig. 1. Annotations in PASCAL VOC [12] format.

The provided training data has 21041 total images. It consists of 2829 images from Czech (CZ); 10506 from Japan (JP); and 7706 from India (IN) with annotations stored in individual XML files. In Fig. 1, We can see the file structure, bounding box in xml tags and corresponding image example.

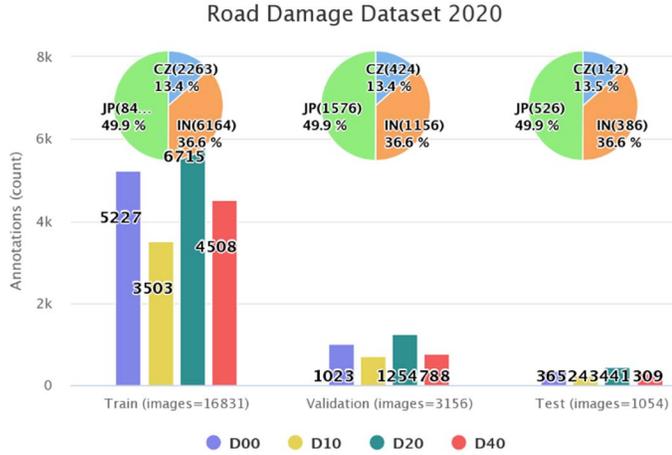

Fig. 2. Train (T), Validation (V) and Test (T) data split for experiments. Bars are for 4 damage class labels D00, D10, D20, D40 provided in the dataset.

The shared Test data are divided into two sets. Test 1 consists of 349 Czech, 969 India and 1313 Japan Road images without annotated ground truth. Test 2 consists of 360 Czech, 990 India and 1314 Japan Road images without annotated ground truth. The detection results on these test images is submitted to the challenge [23] for Avg F1 score evaluation.

In order to run the experiments, we split the given training dataset proportionally into 80:15:5 :: Train (T):Val (V):Test (T) data. This gives us the final image & annotations count in Fig. 2 that will be used for training and tuning.

As we fine tune the models, we need to create composite datasets with Train+Test (T+T) and Train+Val (T+V) dataset composition. This will help model use entire data for learning and evaluation.

### B. Evaluation Strategy

Evaluation strategy includes matching of the predicted class label for the ground truth bounding box and that the predicted bounding box has over 50% Intersection over Union (IoU) in area. Precision and recall are both based on evaluating Intersection over Union (IoU), which is defined as the ratio of the area overlap between predicted and ground-truth bounding boxes by the area of their union.

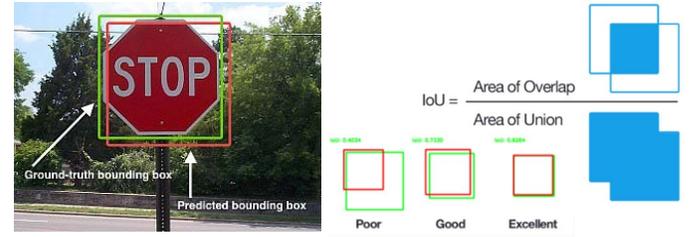

Fig. 3. Illustration for Intersection over Union (IoU) calculation.

The evaluation of the match is done using the Mean F1 Score metric. The F1 score, commonly used in information retrieval, measures accuracy using the statistics of precision p and recall r. Precision is the ratio of true positives (tp) to all predicted positives (tp + fp) while recall is the ratio of true positives to all actual positives (tp + fn). Maximizing the F1-score ensures reasonably high precision and recall.

The F1 score is given by:

$$F_1 = 2 \times \frac{p \times r}{p + r} \ where \ p = \frac{tp}{tp + fp} \ and \ r = \frac{tp}{tp + fn}$$

Avg F1 score serves as a balanced metric for precision and recall. This is the metric we obtain in our private leaderboard, upon submitting the evaluation results on Test 1 or Test 2 datasets.

## III. METHODS

There are two main classes of object detectors that are consistently performing well on the popular Microsoft Common Objects in Context (MS COCO) [6] dataset. In one-stage detection it is YOLO [13], RetinaNet [7] and in two-stage region proposal based Faster R-CNN [4] or Mask R-CNN [8] methods are widely used. Mask R-CNN is an extension of Faster R-CNN with an additional mask proposal branch for segmentation.

YOLO has a single neural network that predicts bounding boxes and class probabilities directly from full images in one evaluation. Since the whole detection pipeline is a single network, it can be optimized end-to-end directly on detection performance.

Faster R-CNN [4] is a region based approaches that predicts detections based on features from a local region. This region is localized using a Region Proposal Network (RPN). The first stage network is for region proposal on the features from convolution backbone and the second stage is a fully connected network for object classification and bounding box regression.

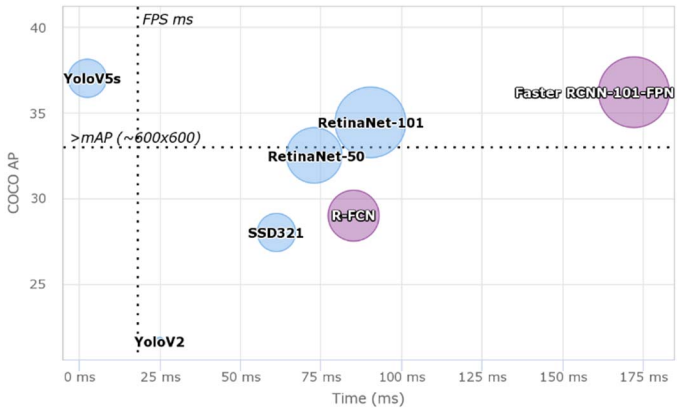

Fig. 4. Illustration of one-stage (blue) and two-stage (pink) detectors time/ accuracy trade-off on MS COCO test data. The size of the bubble represents the relative size of network

Source: Lin et. al. (2017) RetinaNet and Ultralytics Yolo v5

### A. Convolution Backbone

The backbone network is a standard Convolutional Neural Networks (CNN), used to extract high level visual features from the entire image. The high level features are represented as convolutional feature map over the image. Deep Residual Networks [9] like Resnet 50, Resnet 101, ResneXt 101 and Feature Pyramid Network (FPN) or a combination of Resnet and FPN have shown to work well with most object detection models including Faster R-CNN and Mask R-CNN.

One-stage networks like RetinaNet have also used Residual Network and FPN based backbones. YoloV5 on the other hand have made use of Cross Stage Partial Network (CSPNet) [16] to achieve high benchmark on MS COCO datasets.

### B. One Stage Object Detector

Yolo [13] and RetinaNet [7] are popular one-stage object detection models. The accuracy charts are typically lead by two-stage object detectors, whereas one stage detectors are preferred for evaluation speed. One stage detector tends to have low compute requirements and can be easily deployed on smartphone devices.

*Yolo*

You-Only-Look-Once (Yolo) [13] is a unified, real-time object detection algorithm that reformulates the object detection task to a single regression problem.

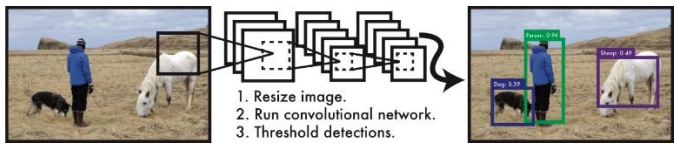

Fig. 5. 1-stage You-Only-Look-Once (YOLO) detector [13]

Yolo employs a single neural network architecture to predict bounding boxes and class probabilities directly from full images. When compared to Faster R-CNN, Yolo provides faster detection with the accuracy trade-off. This has been the core reason for its popularity and multiple extensions and adaptations like Yolov3 [13] and Yolov5 [15] have emerged from it.

Yolov5 [15] includes four different models ranging from the smallest Yolo-v5s with 7.5 million parameters (plain 7 MB and MS COCO pre-trained 14 MB) and 140 layers to the largest Yolo-v5x with 89 million parameters and 284 layers (plain 85 MB and MS COCO pre-trained 170 MB). In the approach considered in this paper, we have experimented with all 4 variants of Yolov5 models. It uses a two-stage detector that consists of a Cross Stage Partial Network (CSPNet) [16] backbone trained on MS COCO [6].

Each Bottleneck CSP unit consists of two convolutional layers with $1 \times 1$ and $3 \times 3$ filters. The backbone incorporates a Spatial Pyramid Pooling network (SSP) [17], which allows for dynamic input image size and is robust against object deformations.

### C. Two Stage Object Detector

Region based CNN (R-CNN) serve as a class of object detection model which falls under two-stage detectors. Faster R-CNN is a region based approach that predicts detections based on features from a proposed region.

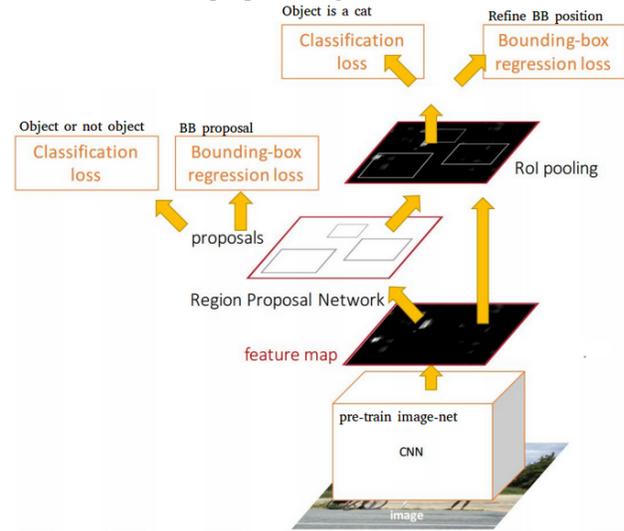

Fig. 6. Illustration of two-stage Faster R-CNN object detector, Ren et. al. (2015). The Region Proposal Network (RPN) is responsible for extracting regions of interest from the feature map directly. Thereafter, a Fully convolutional object classifier is used. The network is shared between RPN and Object detector phase.

*Faster R-CNN*

Region proposal based detectors like Faster R-CNN [4] is a popular two-stage detectors. The first stage generates a sparse set of candidate objects using a Region Pooling Network (RPN), based on shared feature maps, this is classified as foreground or background class. The size of each anchor is configured using hyperparameters. Then, the proposals are used in the region of interest pooling layer (RoI pooling) to generate subfeature maps. The subfeature maps are converted to 4096 dimensional vectors and fed forward into fully connected layers. These layers are then used as a regression network to predict bounding box offsets, with a classification network used to predict the class label of each bounding box proposal.

Feature Pyramid Network (FPN) [5] is employed as the backbone of the network. FPN uses a top-down architecture with lateral connections to build an in-network feature pyramid from a single-scale input. Faster R-CNN with an FPN backbone extracts RoI features from different levels of the feature pyramid according to their scale, but otherwise the rest of the approach is similar to vanilla Resnet. We also employ ResNeXt101 [20] with the FPN feature extraction backbone to extract the features.

## IV. EXPERIMENTS

In this section, we evaluate one-stage Yolo v5 [15] and two-stage Faster R-CNN [4] network with varied pre-processing, backbone network, hyper-parameter tuning and training strategy to achieve better Avg F1 score. We do not use ensemble approach considering that is great for competitions but rarely works well when deployed. In our backbone and methods, we have used Resnet 50, Resnet 101, ResneXt 101 [9, 20] and CSPNet [16] for evaluations considering that when trained these weights can be pruned and compressed to work on smaller devices with minor degradation in accuracy.

In our experiment we start with data pre-processing where we used image augmentation like resize, orientation, and DeepLab V3+ [18] based segmentation to isolate the road surface for downstream assessment.

Next, we look at training a detection model for each country and look at their performance on the submitted F1 score. We also train a single model with the data for all the three countries as a generalized approach. We focus more on the generalized approach considering the theme of this work and the challenge [23] is to obtain a model that can be transferred to other countries.

Finally, we look at thresholding and proposal ranking method applied to the detection results over test datasets. This is important as the output that we submit to the challenge should be the top proposals.

A PyTorch and Detectron2 [14] based framework from Facebook AI Research (FAIR) was used to train and evaluate the Faster R-CNN [4] models while a PyTorch based Yolov5 [15] implementation was used from Ultralytics for comparison purposes. All these implementations are available in open-source Github repository for the community. We were able to customize the data loader and mapper objects to setup the codebase for experimentation. Both these codebases support Tensorboard project for tracking the training accuracy and optimization loss throughout the training process.

The experiments reported in the various tables next, have the model description with epoch runs and chosen backbone network in the first column. Hyper-parameters are described in the second column. Average F1 score is reported for Test 1 and Test 2 dataset based on the kind of experiment we ran.

### A. Pre-processing Images

We looked at segmentation as a way to eliminate background and noise from the image so that we can analyze features only on the road. A PyTorch and Detectron2 [14] based DeepLab V3+ [18] implementation is used for segmentation contours and image cropping.

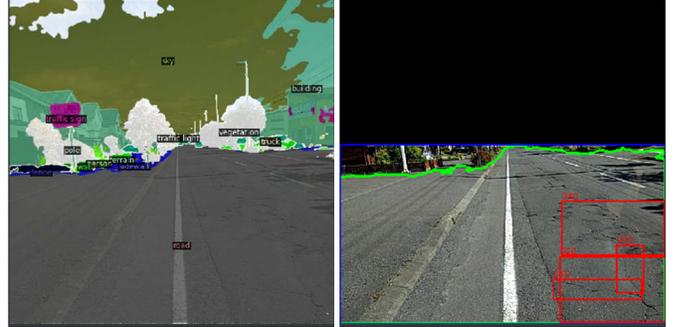

Fig. 7. Illustration of DeepLab based road segmentation. The lower half of the image shows the segment covered by road and the surface damages on it.

We used standard DeepLab V3+ [18] model trained on Cityscape semantic segmentation dataset. The model was able to achieve fair segmentation on most roads in Japan and Czech, while roads in India which had gravel and mud like surface, it did not do a good job of separating the road from surrounding surfaces. We did a basic analysis in Table II to verify, whether segmentation offered an improvement. The dataset used all countries annotation to train a single Faster R-CNN [4] model.

TABLE II. SEGMENTATION BENEFIT

| Model | Hyper-Parameters | Pre-Processing | Avg F1 (Test 1) |
|---|---|---|---|
| Faster R-CNN 27k, Resnet 50 | Batch 128, LR 0.005 | Segmentation | 0.4872 |
| | | None | **0.4945** |

In our experiments, we did not observe any benefit based on our segmentation approach. It appears that the model performance deteriorates and that could stem from segmentation in the India dataset. We proceed without segmentation for the rest of the dataset pre-processing.

### B. Model per country

We trained Faster R-CNN [4] models to fit the data of each country in order to achieve the baseline. The expectation was that the model will achieve better accuracy with three different models dedicated to Czech, Japan and India. We look at a comparison of this approach in Table III.

TABLE III. A MODEL PER COUNTRY COMPARISON

| Model | Hyper-Parameters | Training Approach | Avg F1 (Test 1) |
|---|---|---|---|
| Faster R-CNN 27k, Resnet 50 | Batch 512, LR 0.01 | Single Model for all | 0.5168 |
| | | Multiple Models | **0.5247** |

We get a 1.5% benefit in Average F1 score metric when we train with the baseline Train/Val (T) dataset. However, we take the approach of training a single model across the country's dataset considering the benefits of deployment and model management.

## C. Generalized Model

We attempt to generalize the model by training it on the data from all the countries in the dataset. Here we attempt to compare the two-stage Faster R-CNN [4] and one-stage YoloV5 [15] detection models. We clearly observe in Table IV that two-stage detector out-performs the one-stage detector.

TABLE IV. GENERALIZED MODEL COMPARISON

| Model | Hyper-Parameters | Data | Avg F1 (Test 1) | Avg F1 (Test 2) |
|---|---|---|---|---|
| Faster R-CNN 27k, Resnet 50 | Batch 640, LR 0.01 LR Steps 23k, 25k, 26k | T | 0.5281 | 0.5255 |
| Faster R-CNN 27k, Resnet 50 | Batch 640, LR 0.01 LR Steps 23k, 25k, 26k | T+T | 0.5269 | 0.5270 |
| Faster R-CNN 30k, Resnet 50 | Batch 640, LR 0.01 LR Steps 23k, 25k, 26k | T+V | 0.5289 | **0.5426** |
| Faster R-CNN 30k, Resnet 101 | Batch 4096, LR 0.015 LR Steps 25k, 28k | T+V | **0.5368** | 0.5339 |
| Faster R-CNN 30k, ResneXt 101 | Batch 4096, LR 0.015 LR Steps 25k, 28k | T+V | 0.5255 | 0.5331 |
| YoloV5-s 500, CSPNet | Batch 64 | T | 0.4863 | 0.4701 |
| YoloV5-m 400, CSPNet | Batch 52, Img Size 580 | T | 0.4927 | 0.4714 |

The data used in training these models consists of Train/Val (T) baseline split that is described in the dataset section. We combine Train and Test (T+T) data for training for the second set in the table. Thereafter we improve upon this by composing Train and Val (T+V) data for training the remaining Faster R-CNN model runs. We do gain the expected benefit with this data composition.

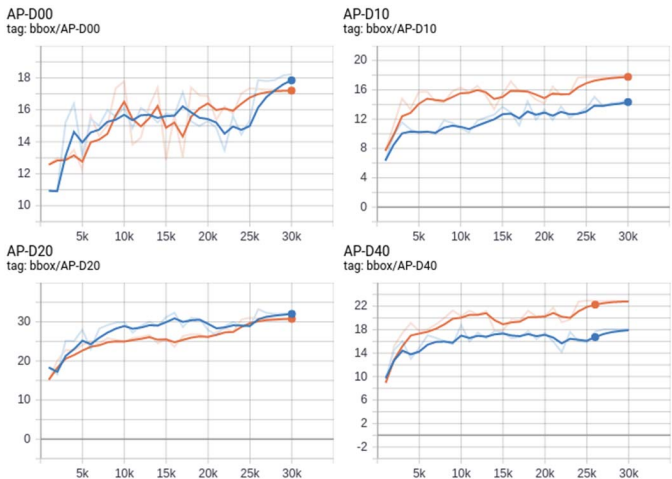

Fig. 8. Average Precision Bounding Box (IoU=.50:.05:.95) for Faster R-CNN Resnet 50_Batch 640 (Orange) and Resnet 101 Batch 4096 (Blue), during the course of training (30k epochs). We observe different accuracies on different damage types like D00, D10, D20 and D40. Due to the sheer weightage of annotations in the dataset, D20 seems to have higher accuracy on both models.

The Model description in Table IV consists of the model name, epoch runs and backbone network. We observe that Faster R-CNN model performs better than YoloV5. The Hyper-parameters includes Batch size, Learning Rate (LR) and LR Step Scheduler. A scheduler decreases the LR by a gamma factor of 0.05 over the steps of mentioned epoch values. LR of 0.01 and 0.015 have performed well with a step schedule of (23k, 25k, 26k) and (25k, 28k) epochs respectively.

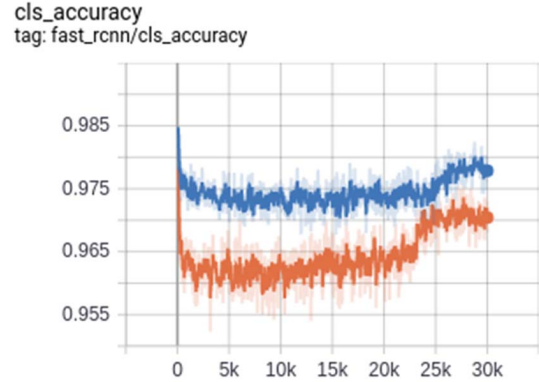

Fig. 9. Classification accuracy for Faster R-CNN Resnet 50 Batch 640 (Orange) and Resnet 101 Batch 4096 (Blue), during training (30k epochs).

We show the best F1 score in Table IV for Faster R-CNN [4] based on a batch size 640 and Resnet 50 [9] in Test 2 evaluation while for Test 1 evaluation score web observe Batch size 4096 and Resnet 101 [9] seems to work well.

We look at the mean accuracy (IoU=.50:.05:.95) on the 5% split Test (T) dataset to monitor and track the progress of the models training. We see in Fig. 8, that this dataset shows high bounding box accuracy on D20 damage type in both the models. However, Resnet 50 with Batch 640 trained model seems to perform well on D10 and D40 damage types, considering both of those classes have relatively low annotations.

When we look at damage classification accuracy in Fig. 9, the Resnet 101 [9] backbone with high batch size demonstrates high accuracies. We also see that the LR step scheduler has a significant impact on the accuracy around 23k for the smaller network and around 25k for the larger network. We also see that the model stops learning around 30k epoch and an early stopping method is used to end the training process. This stops the model from overfitting the training data.

A generalized approach with low network size may allow the model to transfer across countries and reduce the deployment overhead based on the target conditions. However, a bigger network has higher classification accuracy.

## D. Post-processing

In this step we look at operations after detection. The resulting bounding boxes are filtered at 0.7 confidence threshold. Additionally, the detections are sorted by confidence and only the top 5 bounding boxes are sampled for best submission.

## V. RESULTS

We look at all the experiments done and describe the quantitative and qualitative detection results. The quantitative score compares the Average F1 score throughout the experiments. Thereafter, we look at the prediction result overlaid on the images for a visual analysis.

*A. Quantitative Result*

We show an aggregate plot in Fig. 10, with multiple axes for all the experiments run with their Average F1 score and Hyper-Parameters for comparison. This also shows the progress we made as we continued to make changes to the Hyper-Parameters during optimization. The bars are the batch size and black dotted line is the learning rate variations during the experiments.

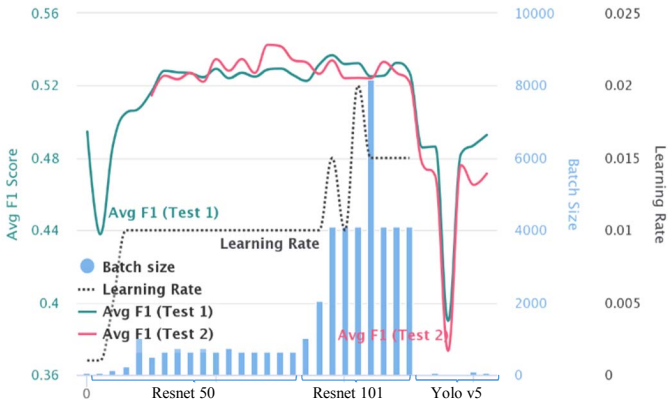

Fig. 10. Comparative result of all experiments Avg F1 score on models.

One observation in Avg F1 score evaluation (green for Test 1, red for Test 2) stood out is that, Resnet 101 [9] with high batch size does better on Test 1 evaluations but does not transfer to better results on Test 2 evaluation. Resnet 50 [9] with an order of magnitude lower batch size performs lower than the Test 1 evaluation of Resnet 101 but evaluates better than Resnet 101 on Test 2 F1 score. The depiction is for a generalized approach, but we have seen that we can always gain approximately 1.5% accuracy benefit by using a per country model approach.

*B. Qualitative Results*

We show the qualitative detection results visually overlaid on the provided Test images. We have used the models trained on (T+V) data composition for showcasing the results. Hence the images selected for this are from the 5% Test (T) data. The model itself is Faster R-CNN [4] with Resnet 50 [9] and FPN [5] backbone network. We will see few good and failed cases here.

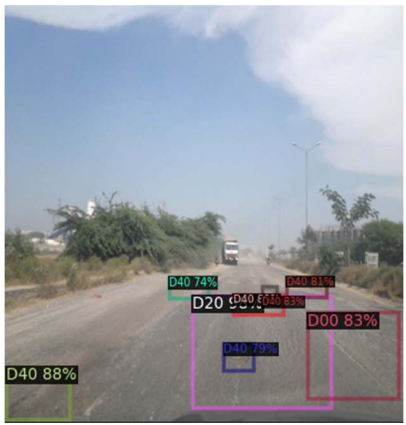

Fig. 11. Successful results of detection on India_000052.jpg Test (T) data using Faster R-CNN [4] Resnet50+FPN [5] model. Trained on data from all countries. Multiple damage types like D20, D00 and D40 have been identified accurately. The model predicts small and medium bounding boxes with more than 70% classification accuracy. The lighting conditions are good and the textures in the captured image is clearly visible.

We show a detection output in Fig. 11, representing the case where all damages have been classified and localized.

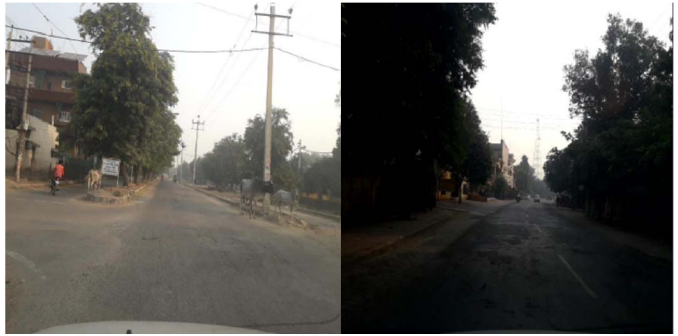

Fig. 12. (left) Failed damage detection results on India_000590.jpg Test (T) data. Unable to detect D00 and D40 damages on right bottom side of the image. (right) Image is captured in a low lighting conditions and has no detections.

The sample in Fig. 12, where the road is in damaged condition. Even though the lighting is good in one of the images, and the network has learnt small or medium damages, this texture spreads through the entire width of the road and may be difficult to learn due to lack of such annotations.

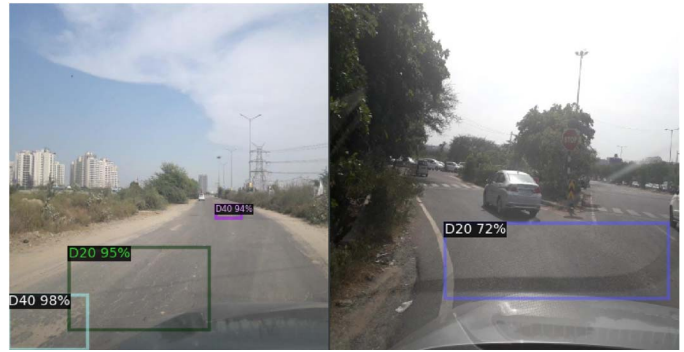

Fig. 13. (left) Damage detection far away on the road surface my not represent the actual damage and raise false positives. (right) Image has dashboard reflection artifacts which is falsely classified as a damage.

*Kind of failures:* In the Fig. 12 & Fig. 13, we have seen images with failed detections. The primary issues in images are due to low light conditions, camera mount positions, artifacts or shadows around objects of interest and inability to geo-fence a spatial range for detections which should be the capture region of interest and avoid looking far down the road.

## VI. CONCLUSIONS

In this work, we experimented with few approaches and went with a single model trained on data across Czech, Japan and India over country specific detection model, due to its simplified deployment process. However, there is an argument against generalization considering training a model per country provides 1.5% improvement over generalized model. Segmentation of road surface as a pre-processing step did not benefit the object detection models considering the ground truth based localized area is not affected by the background features.

We compare the two-stage Faster R-CNN [4] with one-stage Yolov5 [15] detection model for evaluation of framework and observed significant improvement in average F1 score with two-stage network. In our experiments with backbone like FPN [5], Resnet 50 and Resnet 101 [9] in Faster R-CNN [4], Resnet 101 performs and fits the training data and Test 1 evaluation better when compared to its Resnet 50 counterpart which performs better on Test 2 evaluations.

Overall, we see that with larger network and high batch size the model fits the Test 1 evaluation set and otherwise on the Test 2 evaluation dataset. However, the challenge of the unbalanced and limited dataset prohibits the model from learning more. This is where Generative Adversarial Networks (GAN) [21] may be useful or a more structured approach of isolating damage features on road surface using pre-processing. The possibility of learning various parts of the damage may be useful in generation as well as reconstruction of such artifacts in captured images. A semantic segmentation dataset would have benefited by providing the precise structure and texture of the damage. It could have resulted in reducing false positives by better understanding of the damages. Depth information in the image or a spatial fencing in the image may help to limit the detections to damages observed closer to the camera's point of view.